\documentclass[runningheads]{llncs}

\usepackage{eccv}

\usepackage{eccvabbrv}

\usepackage{graphicx}
\usepackage{booktabs}

\usepackage{amsmath}
\usepackage{amssymb}
\usepackage{bbm}

\usepackage{algorithm}
\usepackage{algpseudocode}

\usepackage[accsupp]{axessibility}  

\usepackage{hyperref}

\usepackage{orcidlink}

\newcommand{\MN}{DCPL }

\usepackage[dvipsnames]{xcolor}

\begin{document}

\title{De-Confusing Pseudo-Labels in Source-Free Domain Adaptation}

\titlerunning{De-Confusing Pseudo-Labels in Source-Free Domain Adaptation}



\author{Idit Diamant\inst{1} \and
Amir Rosenfeld\inst{1} \and Idan Achituve\inst{1} \and Jacob Goldberger\inst{2} \and Arnon Netzer\inst{1}}

\institute{Sony Semiconductor Israel \\
\email{\{idit.diamant, amir.rosenfeld, idan.achituve, arnon.netzer\}@sony.com} \and
Faculty of Engineering, Bar-Ilan University, Israel \\
\email{jacob.goldberger@biu.ac.il}}

\authorrunning{I.~Diamant et al.}

\maketitle

\begin{abstract}
Source-free domain adaptation aims to adapt a source-trained model to an unlabeled target domain without access to the source data. It has attracted growing attention in recent years, where existing approaches focus on self-training that usually includes pseudo-labeling techniques. 
In this paper,  we introduce a novel noise-learning approach tailored to address noise distribution in domain adaptation settings and learn to \textbf{de-confuse the pseudo-labels}.
More specifically, we learn a noise transition matrix of the pseudo-labels to capture the label corruption of each class and learn the underlying true label distribution. 
Estimating the noise transition matrix enables a better true class-posterior estimation, resulting in better prediction accuracy.
We demonstrate the effectiveness of our approach when combined with several source-free domain adaptation methods: SHOT, SHOT++, and AaD.
We obtain state-of-the-art results on three domain adaptation datasets: VisDA, DomainNet, and OfficeHome.
\end{abstract}

\section{Introduction}
\label{sec:intro}

Unsupervised domain adaptation (UDA) involves adjusting a predictive model trained on a labeled source domain to perform well on an unlabeled target domain despite possible domain shifts.
Source-free domain adaption (SFDA) imposes an additional constraint that the source domain data are unavailable during adaptation to the target domain. 
Consequently, SFDA depends heavily on unsupervised learning and self-training techniques.

In SFDA, a key obstacle is to reduce the accumulation of errors caused by domain misalignment.
Existing methods in SFDA focus on self-training using target pseudo-labels and entropy minimization techniques.
In \cite{SHOT2020} the effective reduction of error accumulation was achieved by assigning pseudo-labels to the target data using class clusters in the penultimate layer of the model, where the pseudo-labels were determined based on distances from the centroids of the clusters.
However, due to the domain shift, these generated pseudo-labels tend to be noisy. For this reason, several methods have focused on the refinement of the target pseudo-labels during training (e.g., \cite{chen2022self, CSFDA_2023, yi2023sourcefree, zhang2023rethinking}).
One common solution is to update the pseudo-labels at each epoch, thus refining them to better match the target distribution \cite{SHOT2020, zhang2023rethinking}. 
In \cite{zhang2023rethinking}, the authors handle this problem by pseudo-labeling using a strong pre-trained network and filtering out samples with low confidence.

In this work we do not refine the pseudo-labels during target training and do not make any explicit decision on the PLs correctness. 
Instead, we approach SFDA by adopting a learning with label-noise (LLN) perspective and suggest a novel approach tailored for domain adaptation settings to \textbf{D}e-\textbf{C}onfuse the \textbf{P}seudo-\textbf{L}abels. We call our method \textit{DCPL}. 
It involves modifying the established technique of optimizing a noise transition matrix (or a confusion matrix) to adapt the classifier predictions to the pseudo-labels \cite{sukhbaatar2014training, tanno2019learning}. 
Traditional learning-with-label-noise (LLN) methods typically address label noise stemming from errors made by human annotators during labeling. 
In our specific scenario, the label noise is derived from pseudo-labels generated with a specified source model. This label noise emerges from domain shifts, where the characteristics of data in the target domain deviate from those in the source domain used for training. 
In this context, the outputs of the source model carry valuable information about the noise distribution. This information can be leveraged to improve the estimation of the transition matrix. We propose integrating this knowledge into the noise transition matrix learning method to achieve a more accurate prediction of the underlying true label distribution, consequently enhancing the accuracy of classifier predictions.
Empirically, we demonstrate that our approach is easy to integrate with current source-free unsupervised domain adaptation (SFDA) techniques and yields improved prediction accuracy compared to baseline methods, hence achieving new state-of-the-art results.
 Our main contributions are as follows:
\begin{itemize}
    \item We adopt an LLN perspective to approach SFDA using constant noisy labels that do not change during training. 
    \item We introduce DCPL, a novel approach tailored for domain adaptation settings, in which a noise transition matrix is learned to capture the label corruption of the pseudo-labels, thus enabling a better true class-posterior estimation.
    \item We propose integrating knowledge derived from the source model into the noise transition learning to achieve a more accurate prediction of the underlying true label distribution.
    
\end{itemize}

We demonstrate the effectiveness of our approach by implementing it across several SFDA methods: SHOT, SHOT++, and AaD. Our results establish new state-of-the-art performance on three domain adaptation datasets: VisDA, DomainNet, and OfficeHome. Our code can be found at: {\url{https://github.com/ssi-research/DCPL_SFDA}}.

\section{Related Work}

\paragraph{{\bf Unsupervised Domain Adaptation.}}
UDA attracted much attention in recent years \cite{tzeng2014deep, ganin2015unsupervised, ganin2016domain, tzeng2017adversarial, saito2018maximum}. One common approach for addressing UDA involves employing distributional matching (DM). This is usually done by directly minimizing the domain discrepancy statistics \cite{sun2016deep, Morerio2018} or by using some form of domain-adversarial learning \cite{ganin2016domain, tzeng2017adversarial}. However, in \cite{Li2020RethinkingDM} it was shown that DM-based approaches have limited generalization ability under what they term ``realistic shifts". One such shift is when the source label distribution is balanced while the target label distribution is long-tailed. A possible remedy proposed in \cite{Li2020RethinkingDM} is to use self-supervised methods which were found to be useful for UDA \cite{sun2019unsupervised, ge2019mutual, xu2019self, achituve2021self}.

\paragraph{{\bf Source-Free Domain Adaptation.}}
Recently, in light of the growing emphasis on privacy and data protection, there has been a rising interest in source-free domain adaptation \cite{Li_2020_CVPR, SHOT2020,yang2022attracting,A2Net2021,sanqing2022BMD,Qiu2021CPGA, hwang2024sfda}. 
In SFDA, a model is first trained on the source data and then updated based on the target data, usually using unsupervised techniques and self-training due to the absence of labeled data \cite{SHOT2020}. 
Some studies modify the training procedure of the source model to ease the adaptation process by using self-supervision techniques \cite{kundu2022concurrent}, mixup augmentation \cite{kundu2022balancing}, and/or uncertainty quantification \cite{roy2022uncertainty}. 
Another line of research adds unsupervised loss terms to encourage discriminability and diversity in the output during the target adaptation process \cite{SHOT2020, cui2021fast, GSFDA2021, yang2021exploiting, yang2022attracting}. 
For instance, AaD \cite{yang2022attracting} designs the loss such that it encourages the prediction of neighboring samples to be similar and the prediction of non-neighboring samples to be different. This results in simple discriminability and diversity terms. 
In SHOT \cite{SHOT2020} the authors use the conditional entropy and marginal entropy of the predictions as well as cross-entropy with target pseudo-labels. 
Using target pseudo-labels, the authors managed to reduce the error accumulation caused by domain misalignment. SHOT++ \cite{SHOTplus2021} builds on SHOT by adding a rotation-prediction head to allow self-supervised representation learning on the target domain. The samples are then divided into unlabeled and "labeled" based on the prediction confidence, and an additional step of semi-supervised learning is applied \cite{MixMatch2019}.
The authors of \cite{yi2023sourcefree} noticed that the early-time training phenomenon is present in the SFDA settings.
Hence, they added an auxiliary loss to prevent the model from deviating too much from early-time predictions. 

\paragraph{\bf Pseudo Labels for SFDA.} Several SFDA studies have focused on correcting the pseudo-labels of the target samples under the assumption that they are noisy \cite{chen2022self, CSFDA_2023, CoNMix_2023, litrico2023, yi2023sourcefree, ahmed2022cleaning}. 
In \cite{CSFDA_2023} only reliable samples from the target domain were chosen based on prediction statistics. However, this may lead to under-representation of the target domain data distribution. 
In \cite{CoNMix_2023} the authors used label refinement based on similarity statistics between classes across consecutive epochs. However, in cases of high fluctuations in the class assignment, the similarity matrix may be non-diagonally dominant resulting in uninformative or even damaging labels. In \cite{litrico2023} the labels were refined based on the predictions of neighboring examples. However, it requires storing a memory bank of examples which may impose a burden in low-resource environments. Recently, a co-learning strategy  \cite{zhang2023rethinking} was suggested to improve the quality of the generated target pseudo-labels using robust pre-trained networks such as Swin-B \cite{liu2021Swin}. In addition, they refined the pseudo-labels and replaced uncertain ones with the current model output.

\paragraph{\bf Learning with Label-Noise.}
In general, methods for LLN assume either instance-dependent noise or instance-independent noise \cite{song2022learning}. In this study, we assume that the true label is conditionally independent of the features given the observed label. Common approaches for handling label noise with deep learning models include adding regularization \cite{PereyraTCKH17, liu2020early, xia2020robust}, modifying the loss functions and values \cite{liu2015classification, ghosh2017robust, patrini2017making, arazo2019unsupervised}, sample selection \cite{han2018co, jiang2018mentornet, yu2019does, Li_2023_DISC, Tu2023LearningFN}, and architectural changes, such as adding a noise adaptation layer \cite{sukhbaatar2014training, xiao2015learning, Goldberger2017TrainingDN}. Our approach belongs to the last category since it is simple to incorporate into existing learning systems, easily scaled, and tends to work well.
One obstacle to learning a confusion matrix is to prevent converging to degenerate solutions. Various methods have been proposed to address this issue. In \cite{sukhbaatar2014training, tanno2019learning} it was suggested to regularize the matrix trace. The authors of \cite{Goldberger2017TrainingDN} use a joint learning of the network and the confusion matrix with a careful matrix initialization. In \cite{zhang2021learning} it was suggested to maximize the total variation between the prediction vectors of samples. In \cite{Li2021ProvablyEL} the authors proposed controlling the volume of the simplex formed by the columns of the confusion matrix by minimizing its log determinant. The authors of \cite{Lin2023Holistic} advocated a bi-level optimization process of the network's parameters and the confusion matrix. In this study, we chose to minimize the trace as we found it to work well.

\section{SFDA Problem Setup}
\label{section:setup}

This study was designed to address the challenge of source-free unsupervised domain adaptation (SFDA) in visual classification tasks.
Throughout this work, we deal only with a single source domain and a closed set scenario, where both the source and target domains encompass an identical set of classes.
We investigate SFDA within the context of a $K$-way image classification task.
Typically, this problem is formulated as a two-stage training process. 

In the initial stage, we train a predictive function $h_s: \mathcal{X}_s \rightarrow \mathcal{Y}$ using $n_s$ labeled samples $\{x_{s,i}, y_{s,i}\}_{i=1}^{n_s}$ from a source domain $\mathcal{D}_s$ where $x_{s,i} \in \mathcal{X}_s$ and $y_{s,i} \in \mathcal{Y}$ represent the samples and their associated labels respectively. Unless explicitly stated otherwise, we omit the sample indices. Generally, $h_s(x_s)=\mathrm{argmax}(g_s(x_s))$ where $g_s$ is the source model, comprising a feature extractor initialized using a pre-trained ImageNet model, and a task-specific classification head.
In the second stage, we are provided with $n_t$ unlabeled samples $\{x_{t,i}\}_{i=1}^{n_t}$ from a target domain $\mathcal{D}_t$ where $x_t\in \mathcal{X}_t$. We then train a target function $h_t: \mathcal{X}_t \rightarrow \mathcal{Y}$ to predict the labels $\{y_{t,i}\}_{i=1}^{n_t}$ of the target samples, by only utilizing the source function $h_s$ and the target domain data without access to the source data.
We define $h_t(x_t)=\mathrm{argmax}(g_t(x_t))$, analogous to $h_s$.

\begin{figure}[t!]
\centering{\includegraphics[width=0.9\textwidth]{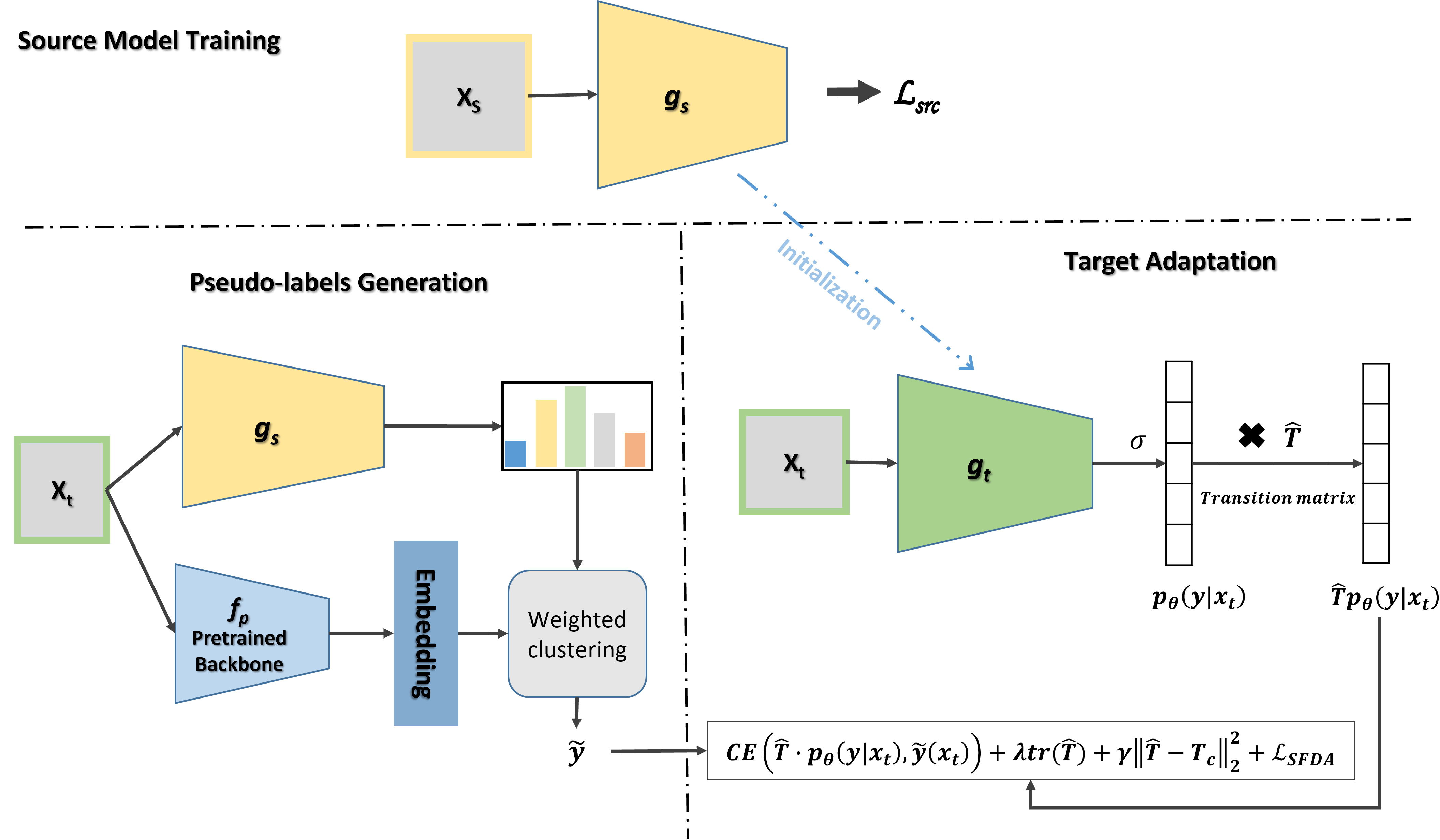}}
\caption{Schematic overview of our framework. DCPL learns both the target model and a transition matrix that adapts the model predictions to noisy pseudo-labels obtained by a general pre-trained network.}
\label{Fig:Framework}
\end{figure}

\section{De-Confusing Pseudo-Labels}
\label{section:method}

Our method is based on the generation of target pseudo-labels, as is commonly done in SFDA approaches.
Due to the domain shift between the source and target domains, these pseudo-labels inherently carry noise. 
Hence, we treat the target adaptation process with these pseudo-labels as a label-noise learning problem, introducing a novel modification to the well-established noise transition matrix LLN method \cite{Goldberger2017TrainingDN, tanno2019learning}.
Conventional LLN methods, like the noise transition matrix approach, generally tackle label noise arising from 
errors made by human annotators during the labeling process, inconsistencies or disagreements among different annotators (inter-annotator variability), mislabeling of instances in the dataset, and/or the presence of ambiguous or subjective data points that are challenging to label accurately. 
However, in our scenario, label noise is introduced by domain shifts, where the characteristics of the data in the target domain differ from those in the source domain used for training.
The label noise is derived from pseudo-labels generated with a specified source model.
Consequently, the distribution of noise depends on the accuracy of the source model on the target dataset, which is associated with the domain gap between the source and target datasets, as well as the generalization capacity of the source model.

In this scenario, the outputs of the source model contain valuable information about the noise distribution that can be utilized to enhance the estimation of the transition matrix. Therefore, we propose integrating this knowledge into the noise transition matrix learning method to achieve a more accurate prediction of the underlying true label distribution and hence enhance the accuracy of classifier predictions.

\subsection{Pseudo-label generation}
\label{par:pl_generation}

At the beginning of the target adaptation process, we first apply a pseudo-labeling function using a pre-trained network to produce high-quality pseudo-labels similar to \cite{zhang2023rethinking}.
There are some advantages to using pre-trained models to produce pseudo-labels. One is that pre-training datasets, such as ImageNet, are large, diverse, and less source-biased and thus may better capture the target distribution. Another advantage is that modern state-of-the-art architectures are designed to learn robust and generalized features, enabling better transfer to target tasks \cite{kim2022unified}. However,
since the pre-training and target label spaces are different, the final classification layer of the pre-trained model is not suitable for the target data. To address this issue, instead of using the pre-trained classifier, the classifier of the source model can be used.

The pseudo-label generation process includes using a feature extractor of a pre-trained robust network (Swin-B) instead of that of the source model.
However, unlike Co-learning \cite{zhang2023rethinking} (and also SHOT \cite{SHOT2020}) which updates the pseudo-labels every epoch, we only invoke the pseudo-labeling function once, at the beginning of the adaptation process. 
The centroid $C^{(k)}$ is thus calculated as the sum of the pre-trained features $f_p$ weighted by estimated class probabilities of the source model $g_s$:

\begin{equation}
\begin{aligned}
    &C^{(k)} = \frac{\sum_{x_t \in \mathcal{X}_t}  \sigma^{(k)}(g_s(x_t))f_p(x_t) } {\sum_{x_t \in \mathcal{X}_t}  \sigma^{(k)}(g_s(x_t))},
\end{aligned}
\label{eq:centroid_with_pretrained_net}
\end{equation}
where $\sigma$ is a softmax function.
Then, for each target instance, a pseudo-label $\tilde{y}$ is generated based on its nearest centroid using the cosine distance.
We denote the final pseudo-labeled target domain $\{x_{t,i}, \tilde{y}_{t,i}\}_{i=1}^{n_t}$ by $\tilde{\mathcal{D}}_t$.

Utilizing robust pre-trained networks like Swin-B enhances the quality of the generated target pseudo-labels through the pseudo-labeling process. Nevertheless, the dependence of these generated pseudo-labels on the source model introduces potential noise, particularly in the presence of domain shifts.

\subsection{Adaptation as learning with noisy pseudo-labels}

After obtaining the pseudo-labels of all the target inputs, we now want to apply a noisy-label learning approach to overcome the label noise. 
In this specific context, our focus revolves around learning techniques designed for handling noisy data by estimating the label noise matrix. These methods have proven highly successful, yielding cutting-edge results \cite{ tanno2019learning, Li2021ProvablyEL}. Our domain adaptation approach relies on the estimated noise matrix to achieve a noise-robust adaptation.
We propose using a modification of this LLN approach to capture the label noise corruption and improve the true label probability estimation. 
Since the noise in our setting originates from the source model, we adjust this approach to the domain adaptation setting by leveraging pseudo-label softmax information and using it to regularize the transition matrix learning. Fig. \ref{Fig:Framework} provides a schematic overview of our overall framework. 

Training methods for noise robustness that involve estimating the noise matrix are all grounded in a fundamental observation. Specifically, the clean class-posterior, denoted as $p(y|x)$, can be inferred by utilizing the noisy class-posterior, represented as $p(\tilde{y}|x)$, and the class-dependent noise matrix $T \in [0,1]^{K\times K}$, where $t_{ij} = p(\tilde{y}=i| y=j)$. This relationship is expressed as follows:
\begin{equation}
\begin{aligned}
    & p(\tilde{y}|x)\, =  T \cdot p(y|x).
\end{aligned}
\label{eq:discrete_probability}
\end{equation}

Various works have incorporated changes to the architecture of classification networks to improve the representation of the label noise matrix in noisy datasets \cite{Li2021ProvablyEL, Goldberger2017TrainingDN, sukhbaatar2014training, tanno2019learning}. These modifications include adding a noise adaptation layer on top of the softmax layer and developing a specialized architecture. The noise adaptation layer is designed to emulate the label transition behavior in learning a network. These adjustments have resulted in improved generalization by adjusting the network output based on the estimated label transition probability.
In our implementation of training with noisy labels, we adhere to the approach outlined in \cite{sukhbaatar2014training, tanno2019learning}.

Denote the softmax label prediction of the adapted network for a sample $x_t$ by ${p_{\theta}}(y|x_t)=\sigma(g_t(x_t))$, where $\theta$ is the parameter set of the network, and the transition matrix estimator by $\hat{T}$ which approximates $T$.
The product of these two represents the estimated class probability of the pseudo-labels $\hat{T} \cdot {p_{\theta}}(y|x_t)$.
During inference, we eliminate the noise adaptation layer defined by $T$ since we want to predict the true (clean) label.
Given the target training inputs $\{x_{t,i}\}_{i=1}^{n_t}$ and the generated noisy pseudo-labels $\{\tilde{y}_{t, i}\}_{i=1}^{n_t}$, 
the standard cross-entropy loss function is:
\begin{equation}
\mathcal{L}_{ce} = 
- \mathbb{E}_{(x_t,\tilde{y}_t)\in \tilde{\mathcal{D}}_t}
\text{CE} \; ( {p_{\theta}}(y|x_t),\tilde{y}_t).
\label{eq:CE-PL-NTM}
\end{equation}

Following  \cite{sukhbaatar2014training, tanno2019learning}, we optimize the network parameters 
alongside the estimated transition matrix  $\hat{T}$ and minimize the average cross-entropy between the pseudo-labels and the estimated noisy label distribution:

\begin{equation}
\begin{aligned}
\mathcal{L}_{ce}=&- \mathbb{E}_{(x_t,\tilde{y}_t)\in \tilde{\mathcal{D}}_t}
\text{CE} \; ( \hat{T} \cdot {p_{\theta}}(y|x_t),\tilde{y}_t) + \lambda \; tr(\hat{T}),
\end{aligned}
\label{eq:CE_with_trace}
\end{equation}
where $tr(\hat{T})$ denotes the trace of matrix $\hat{T}$ and $\lambda$ is a balancing hyper-parameter.

By minimizing this cross-entropy loss, the prediction of the noisy label probabilities, $\hat{T} \cdot {p_{\theta}}(y|x_t)$, is pushed towards the distribution of the generated pseudo-labels $\tilde{y}(x_t)$. However, convergence to the correct solution of $\hat{T}$ and $p_{\theta}(y|x_t)$ is not guaranteed, since there are infinite combinations that can construct a noisy label prediction with a perfect match to the pseudo-label distribution.
To ensure that $\hat{T}$ converges to the true noise transition matrix, namely $\hat{T}\rightarrow T$, a regularization term is added to the loss, thus minimizing the trace of the estimated noise transition matrix.
With this regularization, for $\hat{T}$  to converge to the true noise transition matrix $\hat{T}\rightarrow T$, two assumptions must hold: 1) $\hat{T}$ and $T$ are diagonally dominant, and 2) the full model, namely, the composition of the noise model on the base classifier, aligns with the label distribution of the pseudo-label generator.
Then, it can be shown that the diagonal elements of the estimated transition matrix upper bound the true matrix:
\begin{equation}
\begin{aligned}
    &t_{ii} = \sum_{j}  \hat{t}_{ij} \bar{p}_{ji} \leq \sum_{j} \hat{t}_{ii}\bar{p}_{ji} = \hat{t}_{ii}(\sum_{j}\bar{p}_{ji}) = \hat{t}_{ii},
\end{aligned}
\label{eq:trace_proof}
\end{equation}
where $\bar{p}_{ji} \triangleq \mathbb{E}_{p(x| y = i)}[p(\arg\max\limits_{k}{p_{\theta}}^{(k)}(y|x) = j | y = i)]$.
Moreover, it was shown in \cite{tanno2019learning} (Lemma 1) that obtaining $\hat{T}$ with a minimal trace uniquely coincides with $T$. Note that assumption (2) holds by bringing $\hat{T} \cdot {p_{\theta}}(y|x_t)$ closer to the true noisy distribution $\tilde{y}$, which is encouraged by minimizing the cross-entropy.  As for assumption (1), using a large pre-trained network (Swin-B \cite{liu2021Swin}) and a source model that is trained on a source domain with a reasonable domain gap from the target domain will probably generate a diagonally dominant $T$ (see Section \ref{section:Experiments} for a transition matrix example). 

\paragraph{\bf{Domain adaptation regularization.}} Minimizing the trace as above provides a structural regularization on the transition matrix. However, having a trained model as the source of the pseudo-labels allows us to impose an additional, more powerful prior, specific to the setting of domain adaptation.
More specifically, we leverage the softmax values of the pseudo-labels, positing that they correlate well (on average) with the true confusion matrix of the pseudo-labeler. 
The softmax values of the pseudo-labels are determined by computing the cosine distance to the target centroids.  
Computing the mean softmax vector for each cluster provides insights into class similarities, which can serve as indicators of the pseudo-label noise.
Stacking these vectors forms a prior confusion matrix.
The aim is to leverage this prior to regularize the learning of the transition matrix $\hat{T}$ which is initialized in a class-agnostic manner.

We define this prior $T_c$ on the confusion matrix as follows.
Let $\mathcal{D}^k_t = \{x_t: \tilde{y}_t=k\}$ be the set of all samples pseudo-labeled as class $k$. 
We first compute the pseudo-label logits $s(x_t)$ obtained via cosine-similarity with the centroids calculated with Eq. \eqref{eq:centroid_with_pretrained_net}:

\begin{equation}
\begin{aligned}
    &s(x_t)[k] = \frac{f_p(x_t)\cdot C^{(k)}} {\lVert f_p(x_t)\rVert \lVert C^{(k)} \rVert}. 
\end{aligned}
\label{eq:PL_softmax}
\end{equation}
We then compute the prior matrix $T_c$ with the mean softmax vector for each class $k$ as follows:

\begin{equation}
\begin{aligned}
    &T_c[k,j] = \mathbb{E}_{x_t \in \mathcal{D}^k_t } \frac {\text{exp} ( s(x_t)[j]/\tau)} {\sum _{i=1}^{K}\text{exp} ( s(x_t)[i]/\tau)},
    \end{aligned}
\label{eq:CM_estimate_based_PL_softmax}
\end{equation}
where $\tau$ is a temperature (set to 0.01 unless stated otherwise) that is applied for sharpening the predictions, given that the logits are computed using cosine similarity constrained within the range $[-1,1]$. 
Finally, we add a term to the loss to encourage the learned transition matrix to be similar to the estimated prior:
\begin{equation}
\begin{aligned}
    \mathcal{L}_{DCPL}=& - \mathbb{E}_{(x_t,\tilde{y}_t)\in \tilde{\mathcal{D}}_t}
\text{CE} \; ( \hat{T} \cdot {p_{\theta}}(y|x_t),\tilde{y}_t) + \lambda \; tr(\hat{T}) +\gamma \; \lVert \hat{T}- {T}_{c}  \rVert ^2_2 ,
\end{aligned}
\label{eq:CE_with_trace_and_prior}
\end{equation}
where $\gamma$ is a balancing hyper-parameter.

Equipped with our new cross-entropy loss, defined in Eq. \ref{eq:CE_with_trace_and_prior}, we can use it instead of the standard cross-entropy loss term in existing SFDA frameworks, and retain additional regularization terms specific to the SFDA methods, such as the information-maximization loss in SHOT \cite{SHOT2020} or the discriminability and diversity loss terms in AaD \cite{yang2022attracting}. The final loss is then:
\begin{equation}
\begin{aligned}\mathcal{L}=\mathcal{L}_{DCPL}+\mathcal{L}_{SFDA}.
\end{aligned}
\label{eq:final_loss}
\end{equation}
Algorithm \ref{alg:algorithm1} illustrates the DCPL approach.

\begin{algorithm}[t!]
\caption{DCPL algorithm}\label{alg:cap}
\hspace*{\algorithmicindent} \textbf{Input}: Source model $g_s$, Target dataset $\mathcal{D}_t =\{x_{t,i}\}_{i=1}^{n_t}$, pre-trained feature extractor $f_p$, $N$ - number of epochs, $\mathcal{L}_{SFDA}$ - additional loss terms, $\lambda, \gamma$ - hyper-parameters weighing the regularization terms for the transition matrix  \\
\hspace*{\algorithmicindent} \textbf{Output}: target adaptation model $g_t$
\begin{algorithmic}[1]
\State Initialize weights of adaptation model $g_t$ with $g_s$
\State Calculate target centroids $C^{(k)}$ using Eq. \ref{eq:centroid_with_pretrained_net}
\State Calculate pseudo-labels $\tilde{y_t}$ based on Cosine distance from $C^{(k)}$
\State Compute $T_c$ using Eq. \ref{eq:CM_estimate_based_PL_softmax}
\State Initialize $\hat{T}$ with identity matrix
\For{epoch i=1:N} 
\State Compute loss: $   - \mathbb{E}_{(x_t,\tilde{y}_t)\in \tilde{\mathcal{D}}_t}
\text{CE} \; ( \hat{T} \cdot {p_{\theta}}(y|x_t),\tilde{y}_t) + \lambda \; tr(\hat{T}) +\gamma \; \lVert \hat{T}- {T}_{c} \rVert ^2_2+ \mathcal{L}_{SFDA}$
\State Update $\hat{T}$ and $g_t$ weights

\EndFor
\State \textbf{return} adapted model $g_t$
\end{algorithmic}
\label{alg:algorithm1}
\end{algorithm}

\section{Experiments}
\label{section:Experiments}

In this section, we describe our experimental setup and present the results for three different domain adaptation benchmarks: OfficeHome, VisDA and DomainNet.

\subsection{Datasets}

\textbf{VisDA} \cite{peng2017visda} is a large-scale benchmark for synthetic-to-real adaptation. 
It contains over 200K images from 12 classes.
\textbf{DomainNet} \cite{peng2019moment} is a large UDA benchmark for image classification containing 345 classes in 6 domains. Since its full version suffers from labeling noise, following \cite{zhang2023rethinking} we use a subset that contains 126 classes from 4 domains: Real (R), Clipart (C), Painting (P) and Sketch (S).  
\textbf{OfficeHome} \cite{HashingDA2017} is an image classification benchmark containing 65 categories of objects. 
It consists of four distinct domains: Art (A), Clipart (C), Product (P) and RealWorld images (R).

\begin{table*}[!t]
\caption {Performance comparison for VisDA with a Resnet-101 backbone. DCPL and co-learning results are obtained using SwinB-1K. * indicates results obtained with our implementation.}
\centering
\resizebox{1.0\textwidth}{!}{%
    \begin{tabular}{@{}lccccccccccccc@{}} 
    \toprule
     Method (Synthesis→Real) & Plane & Bcycle & { Bus } &{ Car } &  Horse & Knife & Mcycl & Person & Plant & Sktbrd & Train & Truck & AVG\\
    \midrule
    Source only  & 60.9 & 21.6 & 50.9 & 67.6 & 65.8 & 6.3 & 82.2 & 23.2 & 57.3 & 30.6 & 84.6 & 8.0 & 46.6 \\ 
    A$^2$Net \cite{A2Net2021} & 94.0& 87.8 &85.6& 66.8& 93.7& 95.1& 85.8& 81.2& 91.6 &88.2& 86.5 &56.0& 84.3\\
    G-SFDA \cite{GSFDA2021} &  96.1 &88.3 &85.5& 74.1& 97.1 &95.4& 89.5& 79.4& 95.4 &92.9& 89.1& 42.6& 85.4\\
    SFDA-DE \cite{SFDADE2022}&  95.3 &91.2& 77.5& 72.1 &95.7& 97.8 &85.5& 86.1& 95.5 &93.0& 86.3& 61.6& 86.5\\
    CoWA-JMDS \cite{LeeICML2022}  & 96.2 &89.7& 83.9& 73.8& 96.4& 97.4& 89.3& 86.8 &94.6 &92.1 &88.7 &53.8 & 86.9 \\
    RCHC \cite{diamant2022reconciling} & 96.0 & 90.1 & 85.0 & 73.1 & 95.9 & 97.3 & 87.6 & 84.7 & 93.8 & 91.3 & 86.7 & 51.2 & 86.1 \\
    RCHC++ \cite{diamant2022reconciling} & 97.6 & 88.9 & 88.4 & 84.0 & 97.6 & 97.4 & 92.2 & 86.2 & 97.4 & 92.8 & 92.6 & 41.2 & 87.8 \\
    TPDS \cite{TPDS} & 97.6 & 91.5 & 89.7 & 83.4 & 97.5 & 96.3 & 92.2 & 82.4 & 96.0 & 94.1 & 90.9 & 40.4 & 87.6\\
    Co-learn \cite{zhang2023rethinking} & 99.0 &  90.0 &  84.2 &  81.0  & 98.1 &  97.9 &  94.9 & 80.1  & 94.8  & 95.9  & 94.4 &  48.1 &  88.2\\
    \midrule
    AaD \cite{yang2022attracting}* & 96.8 & 89.9 & 83.9 & 82.7 & 96.7 & 95.7 & 89.5 & 79.8 & 95.5 & 92.6 & 88.8 & 56.5 & 87.4\\
    AaD w/ Co-learn \cite{zhang2023rethinking}* & 97.4 & 90.6 & 84.5 & 83.7 & 97.1 & 97.1 & 91.5 & 84.9 & 96.6 & 94.2 & 90.4 & 57.5 & 88.8\\
    \textbf{AaD w/ \MN (ours) } & 97.9 & 92.7 & 83.8 & 77.9 & 97.4 & 97.3 & 90.6 & 83.7 & 96.6 & 96.5 & 92.0 & 66.6 & \textbf{89.4}\\
    \midrule
    SHOT \cite{SHOT2020} & 94.3 & 88.5 & 80.1 & 57.3 & 93.1 & 94.9 & 80.7 & 80.3 & 91.5 & 89.1 & 86.3 & 58.2 & 82.9\\
    SHOT w/ ELR \cite{yi2023sourcefree} & 95.8 &84.1 &83.3&67.9 &93.9& 97.6& 89.2 &80.1 &90.6 &90.4 &87.2& 48.2 & 84.1\\
    SHOT w/ ELR (w/ SwinB)* & 96.2 & 87.5 & 80.9 & 72.1 & 94.9 & 97.5 & 90.3 & 81.6 & 93.4 & 92.9 & 87.9 & 55.0 & 85.8\\
    SHOT w/ Co-learn \cite{zhang2023rethinking}*& 95.7 & 87.9 & 83.4 & 65.2 & 94.3 & 96.3 & 87.2 & 81.8 & 92.1 & 92.2 & 88.6 & 59.1 & 85.3\\
    \textbf{SHOT w/ \MN (ours) } & 97.0 & 89.1 & 83.5 & 63.3 & 95.5 & 97.7 & 88.6 & 81.1 & 93.9 & 95.3 & 90.0 & 65.7 & \textbf{86.7} \\ 
    \hline
    SHOT++ \cite{SHOTplus2021} & 97.7  &88.4  &90.2  &86.3  &97.9 & 98.6  &92.9 & 84.1 &97.1  &92.2 & 93.6  &28.8 & 87.3\\
    SHOT++ w/ co-learn \cite{zhang2023rethinking}*  & 97.8 & 89.6 & 91.2 & 85.4 & 97.9 & 98.1 & 93.6 & 85.7 & 97.2 & 94.7 & 93.3 & 37.2 & 88.5\\
    \textbf{SHOT++ w/ \MN (ours) } &98.0 & 91.3 & 88.5 & 77.9 & 98.0 & 98.7 & 93.3 & 84.6 & 97.3 & 94.9 & 93.8 & 54.7 & \textbf{89.3}\\
    \bottomrule
    \end{tabular}
   }
\label{tab:visda-performance}
\end{table*}

\begin{table*}[ht!]
 \caption {Performance comparison for DomainNet with a Resnet-50 backbone. DCPL and co-learning results are obtained using SwinB-1K. * indicates results obtained with our implementation.}
 \centering
\resizebox{1.0\textwidth}{!}{%
    \begin{tabular}{@{}lccccccccccccc@{}} 
    \toprule
     Method  &   $C\xrightarrow[]{}P$ & $C\xrightarrow[]{}R$ &  $C\xrightarrow[]{}S$ & $P\xrightarrow[]{}C$ &$P\xrightarrow[]{}R$ & $P\xrightarrow[]{}S$ & $R\xrightarrow[]{}C$ & $R\xrightarrow[]{}P$ &
     $R\xrightarrow[]{}S$ & $S\xrightarrow[]{}C$ &  
     $S\xrightarrow[]{}P$ & $S\xrightarrow[]{}R$ & AVG\\
    \midrule
    Source only  & 49.0 & 62.4 & 51.1 & 48.1 & 72.2 & 43.9 & 58.5 & 64.0 & 49.0 & 60.4 & 53.4 & 61.1 & 56.1\\
    AaD \cite{yang2022attracting}* & 61.1 & 77.4 & 62.2 & 63.3 & 81.5 & 61.3 & 71.0 & 70.0 & 59.8 & 74.8 & 67.2 & 78.8 & 69.0\\
    AaD w/ Co-learn \cite{zhang2023rethinking}* & 66.1 & 81.2 & 64.0 & 66.1 & 83.5 & 62.8 & 72.4 & 71.4 & 62.3 & 75.8 & 70.2 & 82.2 & 71.5\\
    \textbf{AaD w/ \MN (ours)} & 65.8 & 82.1 & 64.6 & 65.2 & 83.9 & 62.7 & 72.7 & 72.3 & 62.0 & 75.5 & 70.2 & 82.2 & \textbf{71.6}\\
    \midrule
    SHOT \cite{SHOT2020} &62.0   & 78.0  &  59.9  &  63.9  &  78.8   & 57.4  &  68.7  &  67.8  &  57.7  &  71.5  &  65.5  &  76.3  &  67.3\\
    SHOT w/ Co-learn \cite{zhang2023rethinking}* & 62.7 & 80.8 & 58.7 & 61.1 & 82.0 & 56.2 & 67.1 & 70.1 & 52.9 & 70.9 & 67.8 & 80.9 & 67.6\\
    \textbf{SHOT w/ \MN (ours)} & 62.9 & 80.6 & 61.7 & 63.3 & 82.1 & 60.7 & 70.1 & 71.0 & 59.8 & 73.4 & 68.7 & 80.7 & \textbf{69.6}\\
    \midrule
    SHOT++ \cite{SHOTplus2021} & 63.0   & 80.4  &  62.5  &  66.9  &  80.0  &  60.0  &  71.2  &  68.7  &  61.2  &  72.8  &  66.6  &  78.1  &  69.3\\
    SHOT++ w/ Co-learn \cite{zhang2023rethinking}* & 63.6 & 81.2 & 61.3 & 64.3 & 82.2 & 59.7 & 68.8 & 70.3 & 57.0 & 73.1 & 68.3 & 81.3 & 69.3\\
    \textbf{SHOT++ w/ \MN (ours)} & 62.7 & 80.7 & 62.8 & 64.7 & 82.5 & 62.3 & 72.0 & 71.6 & 62.3 & 74.6 & 68.6 & 81.1 & \textbf{70.5} \\ 
    
    \bottomrule
    \end{tabular}
   }
  \label{tab:DomainNet-performance}
\end{table*}

\begin{table*}[ht!]
 \caption {Performance comparison for OfficeHome with a Resnet-50 backbone. DCPL and co-learning results are obtained using SwinB-1K. * indicates results obtained with our implementation.}
 \centering
\resizebox{1.0\textwidth}{!}{%
    \begin{tabular}{@{}lccccccccccccc@{}} 
    \toprule
     Method  &  $A\xrightarrow[]{}C$ & $A\xrightarrow[]{}P$ & $A\xrightarrow[]{}R$ & $C\xrightarrow[]{}A$ & $C\xrightarrow[]{}P$ & $C\xrightarrow[]{}R$ & $P\xrightarrow[]{}A$ & $P\xrightarrow[]{}C$ & $P\xrightarrow[]{}R$ & $R\xrightarrow[]{}A$ & $R\xrightarrow[]{}C$ & $R\xrightarrow[]{}P$ & AVG\\
    \midrule
    Source only  & 44.6 & 67.3 &74.8 & 52.7 &62.7 & 64.8 & 53.0 &40.6 & 73.2 & 65.3 & 45.4 & 78.0 & 60.2 \\ 
    A$^2$Net \cite{A2Net2021} & 58.4 &79.0 &82.4& 67.5& 79.3 &78.9 &68.0 &56.2 &82.9 &74.1& 60.5& 85.0 &72.8\\   
    SFDA-DE \cite{SFDADE2022} & 59.7& 79.5& 82.4 &69.7& 78.6& 79.2& 66.1 &57.2 &82.6& 73.9 &60.8& 85.5 &72.9\\
    CoWA-JMDS \cite{LeeICML2022}  & 56.9 & 78.4& 81.0& 69.1& 80.0& 79.9& 67.7& 57.2& 82.4 &72.8 &60.5 &84.5& 72.5 \\
    TPDS \cite{TPDS} & 59.3 & 80.3 & 82.1 & 70.6 & 79.4 & 80.9 & 69.8 & 56.8 & 82.1 & 74.5 & 61.2 & 85.3 & 73.5\\
     Co-learn \cite{zhang2023rethinking}* & 58.1 & 84.4 & 87.6 & 75.9 & 85.5 & 86.9 & 74.6 & 55.1 & 87.9 & 78.0 & 58.2 & 88.4 & 76.7\\
    \midrule
     AaD \cite{yang2022attracting} &59.3 &79.3& 82.1 &68.9& 79.8& 79.5& 67.2 &57.4& 83.1& 72.1 &58.5& 85.4 &72.7\\
     AaD w/ co-learn \cite{zhang2023rethinking}* & 61.6 & 83.6 & 85.5 & 74.6 & 84.5 & 84.2 & 72.3 & 60.7 & 85.7 & 76.8 & 63.0 & 88.0 & 76.7\\
     \textbf{AaD w/ \MN (ours) } & 61.9 & 84.5 & 87.1 & 75.7 & 85.8 & 85.6 & 73.6 & 59.8 & 86.5 & 78.1 & 62.9 & 89.0 & \textbf{77.5}\\
    \midrule
     SHOT \cite{SHOT2020}  & 57.1 & 78.1 & 81.5 & 68.0 & 78.2 & 78.1 & 67.4 & 54.9 & 82.2 & 73.3 & 58.8 & 84.3 & 71.8\\ 
     SHOT w/ ELR \cite{yi2023sourcefree} & 58.7& 78.9& 82.1& 68.5& 79.0& 77.5 &68.2 &57.1& 81.9 &74.2& 59.5& 84.9 & 72.6\\
     SHOT w/ co-learn \cite{zhang2023rethinking}* & 60.8 & 80.8 & 85.5 & 72.4 & 81.9 & 83.3 & 72.4 & 59.4 & 84.7 & 77.1 & 62.3 & 87.3 & 75.7\\ 
     \textbf{SHOT w/ \MN (ours) } & 59.9 & 84.3 & 87.8 & 76.8 & 85.8 & 86.6 & 74.8 & 58.6 & 87.4 & 77.9 & 61.1 & 89.0 & \textbf{77.5}\\
     \midrule
     SHOT++ \cite{SHOTplus2021} & 57.9 & 79.7 & 82.5 & 68.5 & 79.6 & 79.3 & 68.5 & 57.0 & 83.0 & 73.7 & 60.7 & 84.9 & 73.0 \\
     SHOT++ w/ co-learn \cite{zhang2023rethinking}* & 60.1 & 82.2 & 86.1 & 73.2 & 83.3 & 83.7 & 72.2 & 59.6 & 85.1 & 77.0 & 63.2 & 87.9 & 76.1\\
     \textbf{SHOT++ w/ \MN (ours) } & 61.2 & 84.3 & 88.0 & 76.7 & 86.1 & 86.6 & 74.2 & 59.3 & 87.6 & 78.2 & 62.4 & 89.5 & \textbf{77.8} \\
    \bottomrule
    \end{tabular}
    }
  \label{tab:OH-performance}
\end{table*}

\subsection{Implementation Details}
In all our experiments we use PyTorch and train on NVIDIA V100 and A100 GPUs.
We use pre-trained ResNet \cite{he2016deep} architectures as backbones. 
Specifically, Resnet50 is used for OfficeHome and DomainNet, while Resnet-101 is used for VisDA.
As in \cite{SHOT2020,SHOTplus2021}, we replace the original fully connected layer with a bottleneck layer followed by batch normalization and a task-specific fully connected classifier with weight normalization. 
We follow the source model training procedure applied in \cite{SHOT2020,SHOTplus2021}. 
Specifically, we train the whole network through back-propagation, and the newly added layers are trained with a learning rate 10 times larger than that of the pre-trained layers. 
We use mini-batch SGD with a momentum of 0.9 and a weight decay of 1e-3.

For the target adaptation process, we generate the pseudo-labels at the
beginning of training using a pre-trained network Swin-Base \cite{liu2021Swin}, since Swin transformers \cite{liu2021Swin} were shown to be robust under domain shifts \cite{kim2022unified}. In our experiments, we used two pre-trained SwinB networks, one of which was pre-trained on Imagenet-1K, and the other which was pre-trained on Imagenet-22K and then finetuned on Imagenet-1K (ImageNet-22K-1K).
The learning rate is set to 1e-2 for OfficeHome, and 1e-3 for VisDA and DomainNet.
The temperature $\tau$ for calculating $T_c$ is set to 0.01 when using the Imagenet-1K pretrained network and 0.05 for ImageNet-22K-1K. 
$\lambda$ was set to 0.01 \cite{tanno2019learning} for all datasets to avoid overfitting and maintain simplicity for the practitioners.
For $\gamma$ we chose 0.01 for OfficeHome and DomainNet which have a large number of classes and 1 for VisDA with a smaller amount of classes to account for the ratio in class amount (squared due to square in the loss term).
We train the adapted networks for 50 epochs in all cases. 
All of our results report the mean accuracy over three runs with different random seeds (2019, 2020, and 2021) as was done in \cite{SHOT2020}.

\subsection{Main Results}
\label{sec:MainResults}
In this section, we report the results of \MN on all of the datasets. Tables \ref{tab:visda-performance} -- \ref{tab:OH-performance} present performance comparisons of our method to state-of-the-art methods.  
Since \MN only adjusts the cross-entropy loss, it can be combined with other common SFDA methods. Here we investigate its combination with SHOT \cite{SHOT2020}, SHOT++ \cite{SHOTplus2021} and AaD \cite{yang2022attracting}. 
For SHOT and SHOT++ which contain cross-entropy components, we replace this component with our DCPL cross-entropy. For AaD, which originally did not include a pseudo-labeling process or a cross-entropy term loss, we follow co-learning \cite{zhang2023rethinking} and add a cross-entropy term in our comparisons.
Finally, we also report the accuracy of "source only"  which is the source model (trained with the source dataset alone) on the target dataset. 

The tables show the following: (1) \MN outperforms all previous methods on all three datasets, and (2) \MN significantly outperforms Co-learning irrespective of the SFDA method it was applied to.

\begin{table}[t!]
 \caption {Prior loss (Eq. \ref{eq:CE_with_trace_and_prior}) contribution to DCPL performance}
 \centering
\resizebox{0.7\textwidth}{!}{%
    \begin{tabular}{@{}lccccc@{}} 
    \toprule
     Method & { Prior loss } & {OfficeHome } & {DomainNet } & {VisDA }\\
    \midrule
     {SHOT w/ co-learn} & NR & 75.7 & 67.6 & 85.3 \\       
     {SHOT w/ \MN } &  - & {77.4} & 69.3 & 85.9\\
     {SHOT w/ \MN  } & + & 77.5 & 69.6 & 86.7\\
    \bottomrule
    \end{tabular}
    }
  \label{tab:prior-ablation}
\end{table}

\begin{table}[t!]
 \caption {Influence of PL network (SwinB-1K versus SwinB-22K-1K) on performance}
 \centering
\resizebox{0.7\textwidth}{!}{%
    \begin{tabular}{@{}llccc@{}} 
    \toprule
     Method & PL net & {OfficeHome } & {DomainNet } & {VisDA }\\
    \midrule
     {SHOT w/ \MN  } &  SwinB-1K  & 77.5 & 69.6 & 86.7\\
     {SHOT w/ \MN } & {SwinB-22K-1K  }  & {83.8} & {73.7} & {88.0}\\
     \midrule
     {SHOT++ w/ \MN  } & SwinB-1K & 77.8 & 70.5 & 89.2\\
     {SHOT++ w/ \MN } & {SwinB-22K-1K  }  & {83.8} & {74.2} & {89.9}\\
    \bottomrule
    \end{tabular}
    }
  \label{tab:swin-performance}
\end{table}

\subsection{Ablation Study}
\label{sec:ablation}

In this section, we analyze the contribution of the noise transition matrix and the influence of our new prior loss regularization on performance. 

\paragraph{\bf{Prior loss regularization ablation}.} Table \ref{tab:prior-ablation} shows the contribution of our prior loss to the transition matrix learning when applied with SHOT \cite{SHOT2020} implementation for all three datasets. In all cases, the performance increased when the prior was added with the largest contribution in VisDA.

\paragraph{\bf{Pseudo-labeling network ablation.}}
The strength of the pseudo-labeler has a direct effect on the quality of the adapted network, as a stronger network will reduce label noise. To test this, we report results using the SwinB version trained on either ImageNet1K or trained on ImageNet22K and finetuned on ImageNet1K. Table \ref{tab:swin-performance} depicts the performance of both SHOT\cite{SHOT2020} and SHOT++ \cite{SHOTplus2021} with DCPL. 

\paragraph{\bf{Oracle comparison.}}
Table \ref{tab:Ablation-DomainNet} presents a performance comparison of \MN to an oracle implementation on the DomainNet dataset. The oracle version was implemented using the real noise transition matrix based on the target labels (which we do not have access to in UDA). The comparison to the oracle allows us to analyze the performance increase potential of \MN when the estimated transition matrix is made accurate.
In the table, we also compare \MN to a naive baseline implementation in which the transition matrix is set to a fixed identity matrix ($\hat{T} = I$); i.e., if we assume noiseless pseudo-labels and do not learn the TM. 
Comparing this naive baseline to DCPL shows an increase of 4.3\% using SwinB-1K and 2.2\% using SwinB-22K-1K. This demonstrates the contribution of TM over SwinB.
Comparing this naive baseline to the oracle version of \MN shows an increase of 7.2\% to 72.5\% and 5.3\% to 76.8\%, using SwinB-1K and SwinB-22K-1K respectively. As expected, \MN achieves an intermediate accuracy of 69.6\% and 73.7\%. These results show that better estimations of the transition matrix can generate significant improvement gains.
   
\begin{table}[t!]
 \caption {Performance comparison to Oracle and Baseline (without using a transition matrix) for DomainNet.}
 \centering
 \resizebox{1.0\textwidth}{!}{%
    \begin{tabular}{@{}llccccccccccccc@{}} \hline
    \toprule
    Method  & PL net & $C\xrightarrow[]{}P$ & $C\xrightarrow[]{}R$ &  $C\xrightarrow[]{}S$ & $P\xrightarrow[]{}C$ &$P\xrightarrow[]{}R$ & $P\xrightarrow[]{}S$ & $R\xrightarrow[]{}C$ & $R\xrightarrow[]{}P$ &
    $R\xrightarrow[]{}S$ & $S\xrightarrow[]{}C$ &  
    $S\xrightarrow[]{}P$ & $S\xrightarrow[]{}R$ & AVG\\
    \midrule
    SHOT \cite{SHOT2020} & & 62.0   & 78.0  &  59.9  &  63.9  &  78.8   & 57.4  &  68.7  &  67.8  &  57.7  &  71.5  &  65.5  &  76.3  &  67.3\\
    {w/ \MN (w/ $\hat{T}=I$)  } & SwinB-1K  & 63.3 & 79.9 & 53.9 & 57.7 & 81.5 & 52.1 & 61.9 & 68.9 & 49.0 & 66.7 & 67.5 & 80.9 & 65.3\\
    {w/ \MN (ours)}  & SwinB-1K & 62.9 & 80.6 & 61.7 & 63.3 & 82.1 & 60.7 & 70.1 & 71.0 & 59.8 & 73.4 & 68.7 & 80.7 & {69.6}\\
   {w/ \MN Oracle } & SwinB-1K & 64.7 & 83.5 & 64.2 & 67.6 & 83.8 & 64.4 & 73.7 & 72.9 & 65.4 & 76.3 & 70.8 & 83.2 & 72.5\\
   \bottomrule

   {w/ \MN (w/ $\hat{T}=I$ )  } & swinB-22K-1K & 65.8 & 83.5 & 62.1 & 67.4 & 84.8 & 62.1 & 71.1 & 72.9 & 61.2 & 72.2 & 70.1 & 84.8 & 71.5\\  
   {w/ \MN (ours) } & swinB-22K-1K  & 69.2 & 84.6 & 65.3 & 68.9 & 86.2 & 64.9 & 74.1 & 75.0 & 63.3 & 75.5 & 72.9 & 84.7 & {73.7}\\
   {w/ \MN Oracle } & swinB-22K-1K  & 71.7 & 86.8 & 68.6 & 74.6 & 87.7 & 68.8 & 76.1 & 76.3 & 68.4 & 78.7 & 75.5 & 87.8 & {76.8}\\
   \bottomrule
   \end{tabular}
   }
  \label{tab:Ablation-DomainNet}
\end{table}

\begin{figure}[t!]
\centering{\includegraphics[width=0.8\columnwidth]{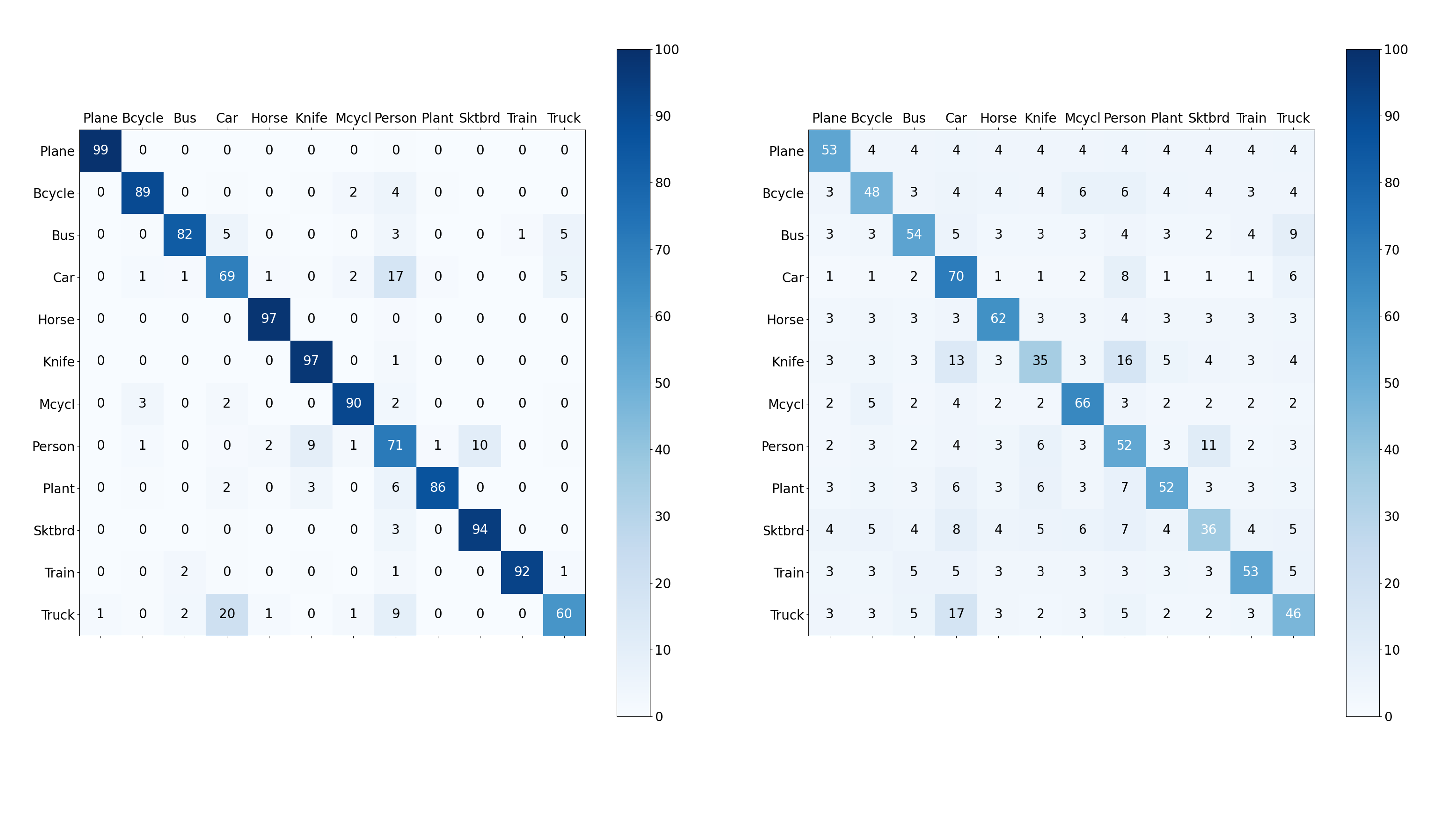}}
\caption{Real and learned noise transition matrices for VisDA.}
\label{Fig:CM}
\end{figure}

\paragraph{\bf{Noise transition matrix.}}
Fig. \ref{Fig:CM} shows an example of real as compared to learned confusion matrices for the VisDA dataset.
It is clear that the real and learned matrices are diagonally dominant, which is compatible with our assumption.
Additionally, the comparison of the matrices shows that the main confusions are learned successfully, e.g., Truck samples classified as Cars, Bus and Car samples classified as Truck, and Person samples classified as Skateboard. On the other hand, there are also estimation errors in the transition matrix, e.g., the Knife category seems to be estimated as a corrupted class, although it has high accuracy estimation (above 90\%) with almost no misclassifications.

\begin{figure*}[t!]
\centering{\includegraphics[width=0.8\textwidth]{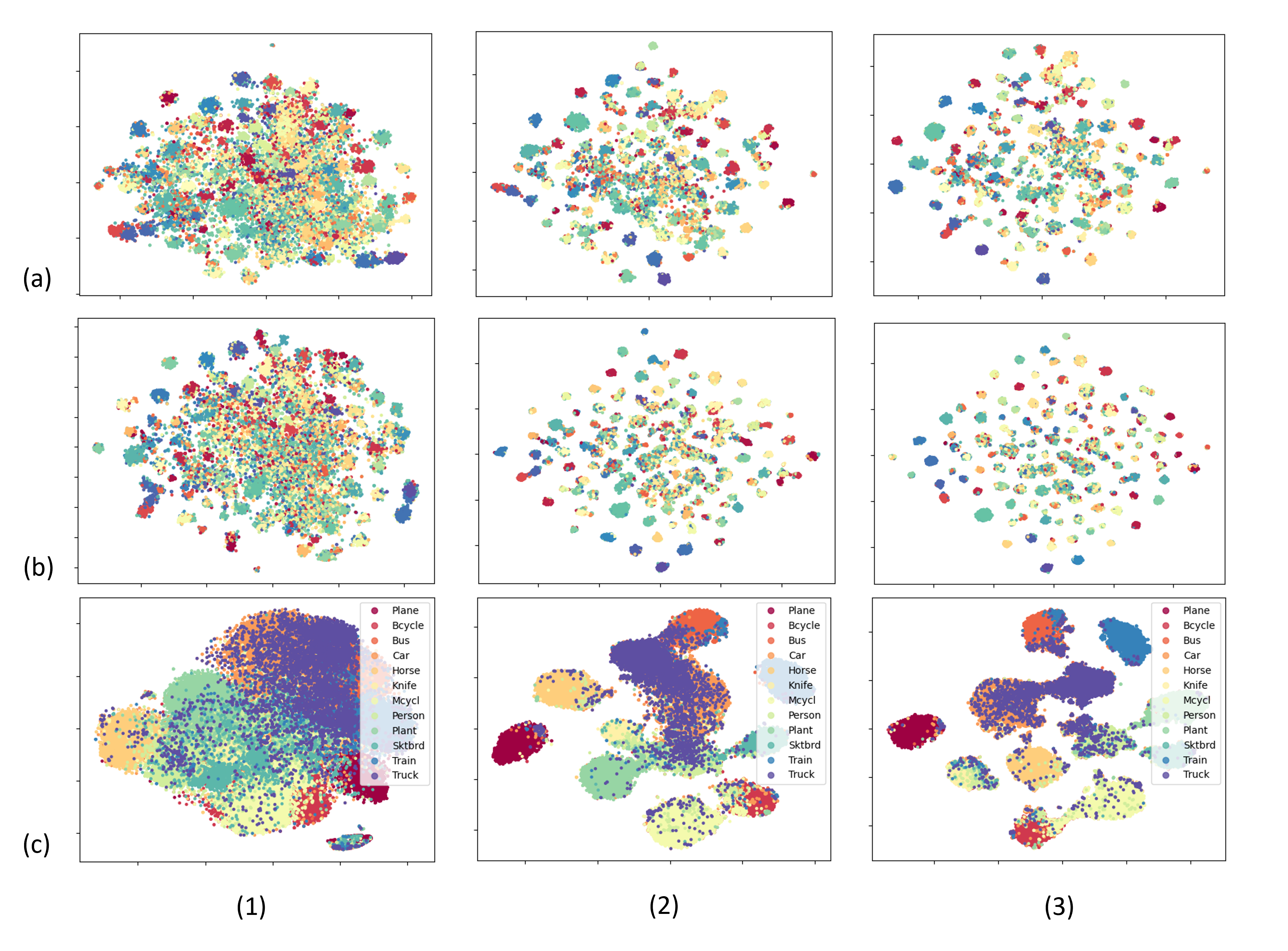}}
\caption{t-SNE visualization for (a)  Clipart to Sketch adaptation from the DomainNet dataset, (b) Real to Sketch adaptation from the DomainNet dataset, (c) synthetic to real adaptation for the VisDA dataset: (1) at the beginning of the target adaptation process, which demonstrates the source model's ability to separate the different classes, (2) at the end of target adaptation using \MN with $\hat{T}=I$, (3) at the end of target adaptation using DCPL.}
\label{Fig:TSNE}
\end{figure*}

\paragraph{\bf{t-SNE visualisations.}}
Fig. \ref{Fig:TSNE} shows t-SNE plots of target sample embeddings using our method for VisDA and DomainNet (Clipart to Sketch and Real to Sketch). These plots show the influence of using the noise transition matrix on the target cluster arrangement. 
In each figure, three t-SNE plots of the target embeddings are presented: (1) at the beginning of the target adaptation process, which demonstrates the ability of the source model to separate the different classes, (2) at the end of target adaptation using \MN with $\hat{T}=I$ (using SHOT), (3) at the end of target adaptation with \MN (using SHOT).
These plots clearly show the contribution of the transition matrix to the discriminability of the different classes.

\section{Conclusion}

In this work, we introduce DCPL, a new SFDA approach that de-confuses the pseudo-labels by learning a noise transition matrix. Inspired by LLN methods and modified for the setting of domain adaptation using information inherent in the source of the pseudo-labels, the transition matrix captures the label corruption, which in turn enables a better true class-posterior estimation.
We demonstrate the effectiveness of our approach when applied to several SFDA methods and provide state-of-the-art results on three domain adaptation datasets, showing that it is complementary to each of the methods we experimented with, as well as applicable to a wide range of scenarios.

%
%
\bibliographystyle{splncs04}
\bibliography{main}
\end{document}